# Generalizable robotic manipulation through compositional skill reuse

Hao Ren[1,2,#], Shuohan Chen[3,#], Senyuan Mu[1], Ruihan Huang[3], Wanfeng Shang[3, 4*], Yajing Shen[1,2*]

1. Department of Electronic and Computer Engineering, The Hong Kong University of Science and Technology, Hong Kong, China

2. HKUST Shenzhen Research Institute, The Hong Kong University of Science and Technology, Shenzhen, China

3. School of Artificial Intelligence, Shenzhen University， Shenzhen, China

4. National Engineering Laboratory for Big Data System Computing Technology, Shenzhen University, Shenzhen, China

# These authors contributed equally.

* Corresponding authors.

## Abstract

Humans can accomplish unfamiliar manipulation tasks by recombining previously acquired skills, without learning each task from scratch. Achieving comparable versatility in robots remains challenging, as data-driven approaches often require extensive action demonstrations, costly training and task-specific policy adaptation. Here we show that complex robotic manipulation can instead be achieved through the composition of reusable skills, without task-specific action-policy training. Inspired by human skill reuse, we introduce a three-layer framework that separates language-based task decomposition, reusable skill composition and execution through fundamental robotic abilities. A language model interprets task goals only once, while each skill draws on existing capabilities for perception, motion planning and closed-loop control, transforming long-horizon manipulation into locally solvable operations. This design eliminates the need for task-specific action demonstrations, avoids continuous large-model inference and enables fast execution and cross-task skill reuse. We demonstrate object sorting, water pouring, plug insertion and package flipping on a real robot, achieving success rates of 93.0% per object, 80.0%, 83.3% and 76.7%, respectively. New tasks can be constructed by recombining existing skills and adding only missing capabilities, reducing deployment time by 40–80% without retraining previously established ones. Our findings suggest that generalizable robotic manipulation need not rely solely on increasingly general action policies, but can emerge from the structured composition and reuse of existing capabilities.

## Introduction

Robotic manipulation remains a fundamental challenge in artificial intelligence and robotics. Despite substantial advances in perception, motion planning and control, robots still struggle to apply existing capabilities to unfamiliar tasks. A robot that can grasp and move objects, for example, may not readily use these abilities to pour water, organize objects or perform assembly operations. Humans, by contrast, can flexibly adapt previously acquired skills to different tasks without learning every operation from scratch. Bridging this gap between acquiring individual capabilities and applying them across tasks remains a major challenge for general-purpose robotic manipulation.

Recent advances in data-driven learning have opened promising avenues towards generalizable robotic manipulation. Imitation learning, reinforcement learning and diffusion-based policies have enabled robots to acquire complex sensorimotor behaviours, while vision–language–action (VLA) models integrate visual understanding, language reasoning and action generation within unified learning frameworks [5–14]. Despite impressive progress, these approaches often require extensive action demonstrations, computationally expensive training and additional adaptation to new tasks. Moreover, learned capabilities are typically encoded within shared model parameters, making individual operations difficult to explicitly isolate, reuse or extend. Repeated inference by large action models can also introduce computational overhead during execution. These limitations raise a fundamental question: must robots learn or adapt a new action policy whenever they encounter a new manipulation task?

Human manipulation offers a different perspective. When faced with a new task, humans can understand the overall goal, organize it into meaningful operations and execute these operations using previously acquired sensorimotor abilities. For example, pouring water involves grasping a container, moving it towards a cup, aligning the opening and controlling the pouring motion—operations that can also be applied in other tasks. This suggests a functional organization across three levels: understanding what needs to be accomplished, selecting and combining appropriate skills, and using underlying sensorimotor abilities to execute them. The deeper implication is that task complexity need not translate directly into learning complexity. Rather than acquiring an entirely new behaviour for each task, existing skills can be recombined to achieve new goals, while unfamiliar operations can build upon previously acquired abilities. Generalizable manipulation may therefore depend not only on how many skills are learned, but also on how effectively they are organized, reused and extended.

Here we present a three-layer compositional framework that translates this principle into robotic manipulation without requiring task-specific action demonstrations or action-policy training. The language-model layer interprets task instructions and produces a structured task specification, which is mapped to executable operations from a shared skill library. The skill layer organizes these operations into a sequence of local goals, while the ability layer provides the perception, geometric reasoning, motion planning and feedback control required for execution. By transforming long-horizon manipulation into locally solvable operations, this architecture separates high-level task reasoning from physical execution, avoiding continuous large-model action inference and enabling efficient execution with limited computational resources. Established skills can be directly reused across tasks, with additional capabilities introduced only when required and without retraining existing skills. We demonstrate the framework on a real robotic arm performing object sorting, water pouring, plug insertion and package flipping, achieving success rates of 93.0% per object, 80.0%, 83.3% and 76.7%, respectively, across randomized spatial configurations. These results support an alternative route towards generalizable robotic manipulation through the structured composition and progressive accumulation of existing capabilities.

## Results

### Decomposing complex manipulation into locally solvable skills

Complex robotic manipulation requires a sequence of decisions involving perception, motion planning, object interaction and feedback control. In end-to-end learning approaches, these behaviours are acquired within a shared action policy that maps observations and task instructions to robot actions (Fig. 1). With $g$ the task instruction, and $o_t$ and $x_t$ the observation and the robot state at time $t$, such approaches can be written as

$$A_t \sim \pi_\theta(\cdot \mid g, o_t, x_t) \tag{1}$$

where $A_t$ is a single action or an action segment and $\theta$ denotes the learned policy parameters. Although such policies can integrate multiple capabilities, learning $\theta$ for the complete task behaviour often requires substantial action data, while evaluating a large $\pi_\theta$ throughout execution can incur repeated inference costs. Importantly, many manipulation tasks contain operations that are already within the robot's capabilities. Learning an entire task policy may therefore be unnecessary when these operations can be identified, executed and reused independently.

We developed a compositional manipulation framework, SHARE, to exploit this structure (Fig. 1). Rather than treating a complex manipulation task as a single sensorimotor problem, SHARE organizes execution into a sequence of skills, each directed towards a defined intermediate goal. A language model $D_\varphi$ decomposes the task instruction $g$ into subtasks $Q$, expressed as a structured task specification, and a compiler $R_\varphi$ maps them to an executable skill program $z$ using a shared skill library $S$:

$$\begin{gathered} Q = D_\varphi(g), \\ z = R_\varphi(Q; S) = [(k_l, \eta_l)]_{l=1}^{L} \end{gathered} \tag{2}$$

where the program $z$ is an ordered sequence of $L$ skills $k_l$ from $S$ with parameters $\eta_l$, such as the target, destination or routing rule. Each skill is implemented through the robot's existing perception, planning and control capabilities, with its own execution conditions, feedback mechanisms and completion criteria. The resulting sequence transforms a long-horizon manipulation task into a series of locally solvable operations, allowing each stage to be executed and evaluated independently.

This organization separates high-level task reasoning from physical execution. Unlike $\pi_\theta$ in equation (1), which is evaluated repeatedly throughout execution, the language model $D_\varphi$ in equation (2) is invoked once to interpret the task and does not generate robot actions. Subsequent operations are handled locally using geometric computation, motion planning and feedback control. For example, a water-pouring task can be organized into approaching the bottle, tracking its position, grasping it, approaching the cup and pouring. Each operation addresses a specific local objective, while the complete sequence achieves the overall task goal. The same approach, tracking and grasping skills can subsequently be invoked in other manipulation tasks without rebuilding their underlying execution mechanisms.

The key consequence of this decomposition is that task complexity no longer has to be addressed entirely through task-level policy learning. Instead, previously established skills provide locally executable solutions that can be combined according to new task requirements. When a task introduces an operation not covered by the existing library, the corresponding skill can be developed and incorporated without modifying unrelated skills. This provides the basis for both efficient task execution and the progressive expansion of robotic manipulation capabilities.

**Composing robotic abilities into reusable skills**

The decomposition of complex manipulation into smaller operations is useful only if each operation can be executed independently and reused under different conditions. To achieve this, SHARE organizes robotic manipulation into three levels: fundamental abilities, skills and tasks (Fig. 2). Fundamental abilities provide general functions such as object localization, state estimation, inverse kinematics, motion planning and feedback control. Rather than developing a separate control policy for every operation, SHARE combines these existing abilities into skills with defined manipulation goals, execution conditions and termination criteria. Each skill generates and adjusts robot actions according to the current state of the environment, allowing the same execution mechanism to be applied across different objects and tasks.

For example, the Approach skill combines visual perception, spatial localization, inverse kinematics and collision-free motion planning to move the robot towards a target. The Track skill incorporates visual feature tracking, target-state estimation and predictive feedback control to continuously correct the relative position between the gripper and a moving object. Pick and Place build on these abilities to grasp, transport and release objects according to their geometry and destination. More specialized skills address distinct interaction requirements. Pour determines the relative poses of a container and a receiving cup before tilting the container to transfer water. Servo Insert combines visual alignment with controlled advancement to insert a plug into a socket, while Flip exploits a planned grasp, lifting motion and timed release to turn over a package. Although these skills perform different operations, they draw on shared perception, geometric computation, motion planning and control capabilities.

A water-pouring task illustrates how these skills are combined into a complete manipulation procedure (Fig. 2). Given the instruction to pour water from a bottle into a cup, the language model produces a structured task specification identifying the operation, target object and destination. The execution layer then assembles the corresponding skill sequence: Approach, Track and Pick to grasp the bottle, followed by Approach and Pour to transfer water into the cup. During execution, the robot determines object positions and manipulation parameters from current observations, while each skill independently generates the motions needed to reach its intermediate goal. The same Approach, Track and Pick skills are also used in other tasks, without modifying their underlying execution logic.

This organization makes skills reusable because their execution mechanisms are separated from the particular objects and tasks to which they are applied. What is reused is therefore the underlying perception, planning and control mechanism, not a fixed motion trajectory: a skill specifies how existing robotic abilities should be coordinated to achieve a local manipulation objective. Changes in object position, geometry or target location are handled through updated sensory information and execution parameters, rather than by relearning the skill. By composing fundamental abilities into closed-loop skills, SHARE provides a common set of executable operations from which different manipulation tasks can be constructed.

**Generalizing manipulation through skill reuse**

A central advantage of compositional manipulation is that skills developed for one task can be reused to accomplish others. Rather than acquiring a separate action policy for each task, SHARE constructs new manipulation procedures by selecting and combining skills from a shared library (Fig. 3). This allows operations that have already been established to retain their execution mechanisms across tasks, while differences in task goals are accommodated through skill selection, sequencing and execution parameters.

We investigated this capability through four manipulation tasks involving different objects, goals and physical interactions. Object sorting established an initial skill library consisting of Approach, Track, Pick and Place. These skills enable the robot to locate and follow objects, grasp them and transport them to designated destinations. When extending the system to water pouring, the existing Approach, Track and Pick skills were directly reused, while only the Pour skill was introduced to perform the additional liquid-transfer operation. The resulting Serve task was executed by combining existing grasping and positioning capabilities with the newly introduced pouring operation, without retraining or modifying the previously established skills.

The same principle was applied to plug insertion and package flipping. For Assemble, the robot reused Approach and introduced Servo Insert to perform visual alignment and controlled insertion. For Flip, Approach and Pick were reused, while a new Flip skill was added to execute the lifting and timed-release motion required to overturn a package. Across all four tasks, Approach was shared by every task, Pick by three tasks and Track by two tasks. Each new task after Sorting therefore required only one additional specialized skill, with the remaining operations drawn from the existing library. This reuse is expected to shorten deployment progressively (Fig. 3): with the setup time of Sorting taken as 100%, the illustrative relative setup time falls to 60% for Serve and 40% for Assemble, so that once the first three tasks are in place, the fourth task, Flip, requires only about 20% of the initial setup time. Importantly, these newly introduced skills were themselves constructed using the robot's existing perception, planning and control abilities, extending the manipulation repertoire without rebuilding its underlying execution capabilities.

This reuse operates at two levels. Fundamental robotic abilities can be combined to construct skills with different manipulation functions, while established skills can be recombined into task sequences with different goals. The introduction of a new task therefore does not necessarily require learning a new action policy or redeveloping previously established operations. Instead, development can focus on the capabilities not yet available in the skill library. Once introduced, these capabilities become reusable components for subsequent tasks, allowing the library to expand incrementally.

These results demonstrate that different manipulation tasks can share substantial portions of their execution capabilities despite requiring different interaction mechanisms. Generalization is achieved not by applying one fixed action sequence to different tasks, but by retaining established skill mechanisms and organizing them according to new task requirements. This provides a practical basis for extending robotic manipulation capabilities through composition and reuse rather than repeated task-specific policy training.

### Adapting reusable skills through sensory feedback

Reusing a skill across different tasks requires more than repeating a previously established motion. Objects may differ in geometry, position and physical properties, and their states may change during manipulation. To accommodate these variations, SHARE separates the execution logic of each skill from the sensory information and parameters that determine its specific actions. As a result, the same skill can operate under different conditions by updating its internal state estimates and execution parameters, without changing its underlying control logic (Fig. 4).

We first investigated this adaptability through visual tracking of moving objects. The Track skill combines object detection, visual feature tracking and three-dimensional state estimation to continuously update the target position and velocity. Rather than directing the robot towards the object's last observed position, the system predicts where the target will be when the motion command takes effect, accounting for perception, computation and actuation delays. We evaluated this mechanism using a bottle and a can moving under different conditions (Fig. 4). As the objects moved, the robot updated its following motion based on the estimated target states, maintaining the spatial alignment required for subsequent grasping. The same tracking mechanism was applied to both objects without modifying the skill's control logic, illustrating how a reusable skill can accommodate changing object positions and motion conditions through feedback.

Sensory feedback also enables skills to respond to physical properties that cannot be reliably determined from object position alone. We investigated this capability in a sorting task involving soft tissue packs and hard blocks. A magnetic tactile sensor integrated into the gripper recorded the contact response during grasping, providing information about object deformation under the applied closure. Four temporal features were extracted from a 0.3-s observation window and used by a lightweight classifier to distinguish soft and hard objects. The measured magnetic responses showed different patterns between the two object categories, and leave-one-run-out cross-validation correctly classified all 52 recorded grasps (Fig. 4). These results demonstrate that contact measurements can provide additional information for selecting subsequent manipulation actions.

Importantly, incorporating tactile information did not require redesigning the overall manipulation procedure. The robot continued to execute the same sequence of Approach, Track, Pick and Place skills. During Pick, the tactile response was used to classify the object, and the resulting category determined the destination for the subsequent Place operation. Thus, new sensory information could influence task execution through the existing skill interfaces, without changing the task sequence or retraining an action policy. Similarly, differences in object position and geometry were handled through updated perception and execution parameters rather than through separate task-specific manipulation policies.

Together, these experiments illustrate how the same skills can adapt to different positions, objects and contact conditions while preserving their execution mechanisms. Visual feedback supports continuous adjustment to target motion, whereas tactile feedback provides information about contact-dependent object properties that can influence subsequent decisions. By separating reusable operation logic from changing sensory states, SHARE enables skill reuse without restricting the robot to fixed action trajectories, providing a basis for adaptation across different manipulation conditions.

### Efficient manipulation without task-specific policy training

We evaluated whether compositional skill reuse enables efficient manipulation across diverse tasks without collecting task-specific action demonstrations or training dedicated action policies. Experiments were conducted on a six-degree-of-freedom robotic arm performing four types of manipulation: object sorting, water pouring, plug insertion and package flipping (Fig. 5). These tasks involve different geometric constraints, contact conditions and operation sequences, providing a test of the framework across distinct manipulation requirements. For each task, we conducted 30 real-robot trials, with target objects and destinations randomly repositioned between trials. All tasks were executed using the shared skill architecture, without collecting robot action demonstrations or training task-specific action policies.

The framework successfully completed all four categories of manipulation. In Sorting, the robot identified objects through tactile feedback and placed them into their corresponding containers, correctly handling 279 of 300 objects across 30 trials (93.0% per object; Wilson 95% confidence interval (CI) 89.5–95.4%). The robot completed all ten objects successfully in 16 of the 30 trials (53.3%; 95% CI 36.1–69.8%). In Serve, the robot grasped a container, aligned it with a cup, poured water and returned the container. Both bottles and cans achieved success rates of 80.0% (12 of 15 trials each; 95% CI 54.8–93.0%), demonstrating the reuse of the same manipulation procedure across two container types. Assemble achieved an 83.3% success rate (25 of 30 trials; 95% CI 66.4–92.7%) through visually guided plug insertion, while Flip achieved 76.7% (23 of 30 trials; 95% CI 59.1–88.2%) by grasping, lifting and releasing a package to produce the desired orientation. These results demonstrate the feasibility of constructing distinct manipulation procedures from shared robotic capabilities and reusable skills.

We next measured execution speed from instruction input to task completion (Fig. 5). Sorting ten objects took 329.07 ± 17.62 s per trial (mean ± s.d., n = 30), or approximately 33 s per object. Successful Serve trials took 46.33 ± 3.34 s with bottles and 49.00 ± 2.63 s with cans (n = 12 each), while successful Assemble and Flip trials took 37.66 ± 4.20 s (n = 25) and 25.33 ± 2.03 s (n = 23), respectively. Across Sorting, Serve, Assemble and Flip, motion planning accounted for only 2.86%, 2.17%, 0.01% and 6.51% of the measured task duration, respectively. The largest measured contribution came from physical action execution, which accounted for 50.26%, 51.24%, 76.45% and 54.27%, respectively. Task duration was therefore limited mainly by the execution of the robot hardware rather than by planning computation.

The framework also enabled execution with limited computational resources. A language model was invoked only once at the beginning of each task to produce a structured task specification, while subsequent perception, motion planning and control were performed locally on a laptop equipped with an NVIDIA RTX 4060 GPU and 16 GB of system memory. This separation between high-level task interpretation and local physical execution avoids continuous large-model action inference and allows the computational demands of execution to be handled by specialized local modules.

The framework required no task-specific robot action demonstrations or action-policy training for any of the evaluated tasks. Instead, existing perception, planning and control abilities were combined into executable skills, and previously established skills were retained when new tasks were introduced. The only trained components were perception modules, including object detectors fine-tuned on labelled scene images and a 25-parameter tactile classifier trained on 52 recorded grasps. For Sorting, for example, the wrist-camera detector was fine-tuned for 25 epochs on 517 training images using the same laptop GPU. By contrast, end-to-end approaches typically rely on robot action demonstrations and on training or adapting an action policy for new tasks [5–14]. SHARE therefore shifts the training requirement from robot action data to perception data: task deployment did not require learning an end-to-end action policy, and new operations could be incorporated without retraining previously established skills.

Together, these experiments demonstrate that diverse manipulation tasks can be executed through the composition of reusable skills without task-specific action-policy learning. The framework combines high-level task interpretation with local perception, planning and control, while allowing previously established capabilities to be retained across tasks. These results support compositional skill reuse as a practical approach to developing efficient and extensible robotic manipulation systems.

## Discussion

Our results suggest that generalizable robotic manipulation need not depend exclusively on learning increasingly comprehensive action policies. Inspired by the compositional nature of human manipulation, we demonstrate an alternative approach in which complex tasks are accomplished by organizing and reusing capabilities that a robot already possesses. Rather than learning an entire sensorimotor policy for each new task, our framework decomposes long-horizon manipulation into locally executable skills that can be combined to achieve different goals. Across object sorting, water pouring, plug insertion and package flipping, this approach enabled diverse manipulation tasks without task-specific action demonstrations or action-policy training. These findings highlight a complementary route towards general-purpose manipulation, in which the ability to recombine existing skills is as important as acquiring new ones.

A key advantage of this approach is that task complexity need not translate directly into learning complexity. By separating high-level task reasoning from physical execution, complex manipulation can be addressed through a sequence of locally solvable operations, each supported by existing perception, planning and control capabilities. This organization avoids continuous large-model action inference and allows established skills to operate under different sensory states and task conditions without retraining their execution mechanisms. More importantly, new tasks can build upon previously acquired skills, with additional capabilities introduced only when necessary. Such incremental expansion enables manipulation capabilities to accumulate over time, rather than requiring complete task policies to be repeatedly developed.

The framework nevertheless has limitations. Its effectiveness depends on whether a task can be decomposed into operations with well-defined intermediate goals and whether the required robotic capabilities are available. Tasks involving complex contact dynamics, deformable objects or strongly coupled interactions may require more sophisticated perception, state estimation and control. In our experiments, pouring was sensitive to liquid-transfer dynamics, insertion could fail when visual features moved outside the camera's field of view, and package flipping depended on grasp stability and release timing. Moreover, new skills currently require manual design and validation, and the system does not yet support high-level replanning in response to execution failures. These limitations constrain its applicability to tasks whose operation sequences and execution conditions can be specified in advance.

Future developments could integrate learning-based methods into the compositional framework, allowing difficult-to-model operations to be acquired as additional reusable skills. Learned contact-control policies, adaptive skill transitions and feedback-driven task replanning could further extend the range of manipulation tasks while preserving existing capabilities. Rather than replacing established perception, planning and control mechanisms, learning could focus on the operations that cannot yet be reliably executed. The generality of robotic manipulation may therefore lie not only in learning more capable policies, but also in making better use of the capabilities robots already possess.

# Methods

## Robot platform, deployment and calibration

Experiments used a reBot Arm B601-RS six-axis robotic arm (Seeed Studio) [44]. Sorting, Serve and Flip used a parallel gripper (maximum opening 89.5 mm), and Assemble used a two-pin tool fixed to the front of the gripper, with its corresponding tool frame. Two Intel RealSense D435-series RGB-D cameras captured at 30 Hz with depth aligned to RGB: camera A was fixed above and in front of the workspace to provide a global view, and camera B was mounted on the wrist to provide a close-up view of the target. A three-axis magnetic tactile sensor (MLX90393) was mounted on the gripper finger, with its normal channel $B_z$ sampled at approximately 160 Hz.

Perception, planning and control all ran on a laptop with an NVIDIA RTX 4060 GPU and 16 GB of system memory, with ROS 2 Jazzy [45], MoveIt 2 [25] (including MoveIt Servo 2.12.4) and Python 3.12 as the software environment. The arm driver sent joint commands at 250 Hz and published joint states at 60 Hz. The two camera streams, the object detectors, three motion-planning processes and the trajectory executor ran as persistent processes, so that no models needed to be reloaded and no planners restarted when a new task started.

The pose of the wrist camera relative to the tool was determined by hand–eye calibration: a ChArUco board [46] was observed from multiple robot poses, the board pose was estimated with the OpenCV iterative PnP method and, combined with the robot forward kinematics, the camera-to-tool rigid transform was solved with Park's method [47]. Calibration used 42 poses for fitting and 14 poses for validation. Camera A was registered to the robot base frame using the same board and the calibrated wrist camera, with 15 frames for fitting and 12 frames for validation. All spatial computation was performed in the robot base frame; the pose of the wrist camera at exposure time was obtained by composing the robot pose at that time with the hand–eye extrinsics.

## Language-based task specification

$D_\varphi$ is implemented by a language model whose role is that of a semantic compiler: it specifies only what the robot should do, not how to do it. The system prompt forbids it from outputting joints, poses, trajectories, waypoints, inverse-kinematics or planner settings, controllers, gripper openings, speeds and safety clearances, all of which are determined at runtime. The language model is called through the structured-output interface of the Kimi K3 API [3], with reasoning effort set to low and a reply limit of 2048 tokens; it is called once, online, in each trial.

The model's reply contains three fields: status, task_spec and clarification. When status is ready, task_spec gives the task specification; when the action, target, destination or quantity is ambiguous, status is needs_clarification, and the model asks a clarifying question through clarification instead of guessing. The action type of a task specification is pick_place or pick_pour_return. The former describes single-destination tasks with target_label or target_labels, target_cardinality and destination_label; when multiple object categories or multiple destinations are involved, routes are instead listed in routes, each route containing a target category, a count and a destination selector composed of a container category plus a spatial relation (for example, the box to the left of the arm, determined as the one with the largest lateral coordinate in the base frame) or a colour. The latter describes a single "pick, pour, return" sequence. Target categories are short lowercase English names that correspond to the detector vocabulary (for example, the Chinese phrase for "blue tissue pack" maps to the detector label blue tissue pack), so that $Q$ can be localised directly by the perception module.

After label normalisation and schema validation, the task specification is compiled by $R_\varphi$ into the skill program $z$; execution policies, such as collision models and the way objects are dropped into boxes, are supplied by the runtime. Because $D_\varphi$ needs to output only a short structured specification, any current mainstream large language model that supports structured output can fulfil this role.

**Skill definition and execution**

Following the options framework [22], each skill is defined as a triple:

$$k = (\mathcal{I}_k, \mu_k, \beta_k) \tag{3}$$

where $\mathcal{I}_k$ is the initiation condition of the skill, $\mu_k(o_t, x_t; \eta)$ is a closed-loop control law and $\beta_k$ is the termination condition. The control law is composed of fundamental capabilities $b \in B$; the fundamental capabilities perform task-independent computations, including object positions in the robot frame, target states, candidate poses, joint solutions, collision-free trajectories and object properties sensed through contact. The termination condition of each skill corresponds to a subgoal $y_\ell$, such as an observation pose, a relative grasp pose or a pouring pose: when the state reaches $y_\ell$, $\beta_{k_\ell}$ is satisfied and execution passes to the next skill. The executor $E$ runs the skill program $z$ (equation (2)) sequentially, taking the state from $x_0$ through $y_1$, $y_2$, ..., to the task goal $x_g$:

$$\begin{aligned} A_t &= E(z, o_t, x_t; S, B) \\ &= \mu_{k_{\ell(t)}} \left(o_t, x_t; \eta_{\ell(t)}\right) \end{aligned} \tag{4}$$

where $\ell(t)$ indexes the skill being executed at time $t$. Parameters can also be determined during execution; for example, in tactile sorting, the destination of each object is determined by the contact response measured after grasping.

**Visual grounding and target state estimation**

**Object detection:** Each task uses two YOLO models (Ultralytics) [23, 24]: an instance-segmentation model for camera A and a detection model for camera B. Both start from public pretrained weights and are fine-tuned on task-scene images captured by the system's cameras and manually annotated. For Sorting, for example, the training, validation and test sets of the wrist model contain 517, 74 and 72 images, respectively, and the model was trained for 25 epochs on the laptop's RTX 4060 GPU.

**3D localization:** For each detection, a depth patch is taken within the object mask, and the median depth and its median absolute deviation (MAD) are computed. The resulting point is back-projected using the camera intrinsics and then transformed into the robot base frame using the camera pose at exposure time, so that the motion of the wrist camera itself is not mistaken for target motion.

**Cross-camera association:** Camera A selects the target instance and predicts the physical support point that camera B should observe. After the arm reaches the observation pose, the system binds a candidate detection from camera B to this prediction using RGB-D support information from the same frame and a gated association test, and initialises local tracking with a mask that separates the target from the table.

**Feature tracking:** The system selects Shi–Tomasi corners [28] within the target mask (quality threshold 0.01, minimum distance 7 px), tracks them frame by frame with pyramidal Lucas–Kanade optical flow [27] (KLT; 21 × 21 px window, three pyramid levels) and filters features with a forward–backward error threshold [48] of 1.5 px. Detection and matching are handled by a separate asynchronous thread and run every other frame; local tracking always reads the latest image, and slower detection results are associated with the tracking record of the same target according to their original exposure time.

**State estimation and prediction:** Target motion is estimated in the base frame by a six-dimensional constant-velocity Kalman filter [49] whose state comprises position $\hat{p}$ and velocity $\hat{v}$, with a white-noise acceleration model for the process noise. The measurement noise is lower-bounded by the noise level of a stationary target and is inflated as the depth MAD increases and as the detection confidence decreases. Measurements whose Mahalanobis innovation exceeds the 95% quantile of the $\chi^2$ distribution with three degrees of freedom, or whose age exceeds 0.25 s, are discarded. A control command takes effect only after image processing has completed and the arm has responded. The controller receives the predicted position rather than the last observed position.

**Pose generation, motion planning and feedback control**

**Candidate poses and inverse kinematics:** Each skill generates a set of candidate end-effector poses from the object geometry, gripper orientation and task constraints, for example a top-down grasp at the centre of the top face of the object's oriented bounding box, or, in the Flip task, a grasp perpendicular to the corner cross-section. Inverse kinematics is seeded with the current joint configuration and solved by three MoveIt solvers (KDL, TRAC-IK [50] and pick_ik), each running in its own persistent planning process.

**Planning and execution:** Free-space motion is planned with OMPL RRTConnect [26, 51] through MoveIt 2, with a collision scene containing the table, the containers and the robot model. Candidates are evaluated in order, and the first candidate that passes the inverse-kinematics, collision and trajectory-limit checks is executed. Straight-line approach and retreat use Cartesian paths with a sampling interval of 2 mm. In multi-object tasks, planning for the next object proceeds in parallel with execution for the current object. Trajectories are first shaped with joint velocity and acceleration limits of 1 rad $s^{-1}$ and 1 rad $s^{-2}$, and are then time-scaled by a factor of 3 when sent through the FollowJointTrajectory interface. After placement, the arm retreats by reversing the trajectory actually executed during descent.

**Tactile sensing and object-property classification**

**Signal acquisition:** The gripper closes at a constant commanded speed of 1.25 rad $s^{-1}$, with no more than 0.005 rad per step. The closing endpoint for the two object types is set according to the visual category: a nominal opening of 38 mm for blocks, whereas a partial opening is retained for tissue packs. After closure, the system records the change in the normal magnetic field, $\Delta B_z$ (μT), over a 0.3 s observation window using the host monotonic clock, with the window starting at the logged event marking the start of the observation stage. Vision-only tasks do not include this observation window.

**Features and classifier:** Without filtering or amplitude normalisation, four temporal features are computed directly from each window: the slope of an ordinary least-squares fit with an intercept; the difference between the medians of $\Delta B_z$ over the final and initial 0.1 s; the root mean square of the linear-fit residuals; and the coefficient of the $t^2$ term of a quadratic fit. After scaling by the per-dimension median and interquartile range of the training data, the features are fed to a multilayer perceptron (4 inputs, 4 tanh hidden units, 1 output; 25 parameters in total), whose output is mapped through a sigmoid and thresholded at 0.5 to classify the object as soft (tissue pack) or hard (block).

**Training:** The training data comprised 52 grasps from 6 acquisition runs, with labels taken from the object category recorded at acquisition. Leave-one-run-out cross-validation was performed with random seeds 42, 43 and 44, with the scaling parameters and the network fitted on the remaining runs in each fold; all 52 grasps were classified correctly. The deployed model was trained on all 52 grasps with seed 42.

**Skills and task composition**

Each skill is defined by five items: its inputs, the fundamental capabilities it calls, its parameters, its termination condition $\beta_k$ and its behaviour on failure. When a skill fails, the task stops and reports the cause.

**Shared skills:** *Approach* calls spatial localisation, candidate poses, inverse kinematics and path planning to bring the arm to an observation pose from which the wrist camera can see the target; it terminates once trajectory execution is complete. *Track* calls target state estimation and following control to maintain target correspondence in the wrist view and to reduce the relative-position error; it terminates once the error has stabilised within tolerance. *Pick* performs a Cartesian descent based on the local target geometry, closes the gripper to the category-specific endpoint and lifts; in tactile sorting, it also records the 0.3 s observation window. *Place* transports the object to the destination and releases it: for a box, the object is released above the centre of the top face of the box's bounding box and falls freely, after which the arm retreats along the reversed descent trajectory.

**Task-specific skills:** *Pour* localises the cup opening with the wrist camera, combines the container geometry with the grasp relation to compute the target position of the bottle mouth above the cup opening and the axial direction of the bottle body, tilts the bottle to the target tilt angle in candidate steps of 5°, and, once pouring is complete, rights the bottle and returns it. *Servo Insert* uses the fixed two-pin tool: after alignment has been completed outside the socket, visual servoing [52] on wrist RGB images at 10 Hz corrects the alignment error in the socket plane (proportional gain $0.6\ s^{-1}$) while the tool advances along the nominal socket normal at 3 $mm\ s^{-1}$ (translational speed cap 8 $mm\ s^{-1}$); the target and maximum depths are both 16 mm, with a timeout of 120 s. *Flip* first uses camera A to select the package corner closest to the arm, and then uses a single wrist observation to determine the contact points on the two edges adjacent to that corner and freezes the reference; the gripper grasps perpendicular to a 20 mm-wide cross-section of the corner (with a preferred downward pitch of 40° or 50°), lifts by 450 mm, hovers for 3 s and then performs a rapid 200 mm motion, opening the gripper when the motion reaches 100 mm, so that the package flips under inertia and lands on the table.

**Task programs:** Sorting executes Approach, Track, Pick and Place in turn for each object; the destination of each object is determined by the tactile classifier after Pick, with soft objects placed in the box to the left of the arm and hard objects in the box to the right, left and right being determined by lateral position in the base frame. Serve executes Approach, Track and Pick on the bottle in turn, then Approach to the cup, executes Pour and returns the bottle; bottles and cans use the same program, with object geometry and parameters supplied by observations and configuration. Assemble executes Approach and Servo Insert. Flip executes Approach, Pick at the package corner and Flip; placement is completed by the package falling freely after release.

**Experimental procedure and success criteria**

Trial counts and object conditions are given in Results and Fig. 5. Each trial started from the same initial configuration and was initiated by a single natural-language instruction, with the positions of the target objects and destinations randomly changed between trials. The success criteria for each task are as follows, with physical outcomes recorded by an operator from on-site observation: for Sorting, per object, the object is correctly classified and placed in the corresponding box; for Serve, water is poured into the cup without spillage and the container is returned; for Assemble, the plug is fully inserted into the socket, bottoming out; and for Flip, the package is flipped and ends with its label facing up. Task duration was measured from instruction input to task completion, including language-model parsing, perception, planning, approach and execution. Success-rate statistics retained all attempts, including failures caused by dropped camera frames or disconnection due to overheating. We report descriptive statistics; metric definitions and sample sizes are given in the corresponding figure legends. Success rates are reported with Wilson score 95% confidence intervals. No hypothesis testing was performed, and no data were excluded.

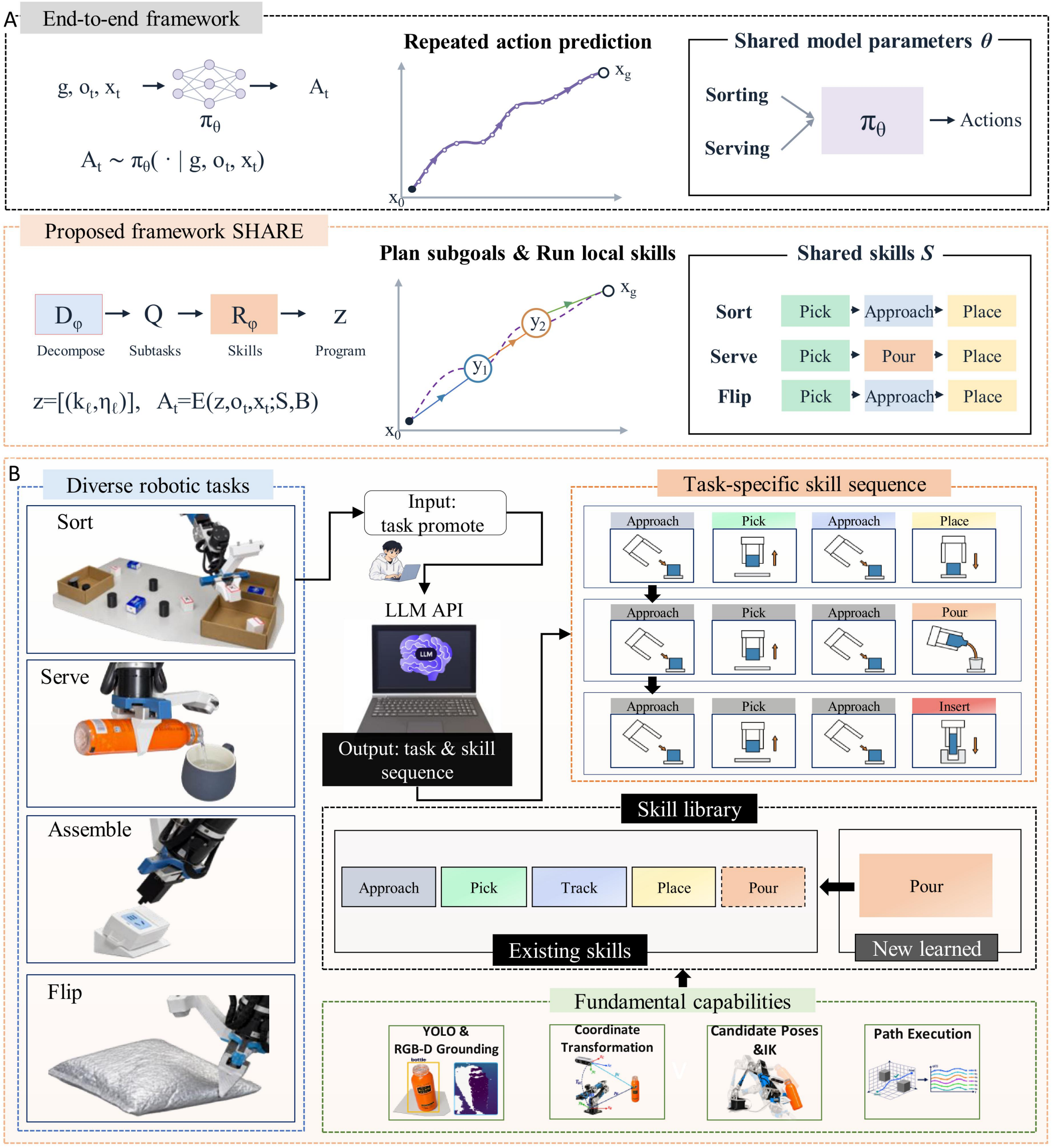

**Fig. 1 SHARE replaces a single end-to-end policy with a program of reusable skills. A**, An end-to-end policy $\pi_\theta$ maps instruction, observation and state directly to actions along the whole trajectory (top); SHARE decomposes the instruction into a skill program $z$ whose local skills reach the goal through subgoals (bottom). **B**, Implementation: a language request is parsed once into a skill sequence drawn from a shared library and executed by fundamental capabilities.

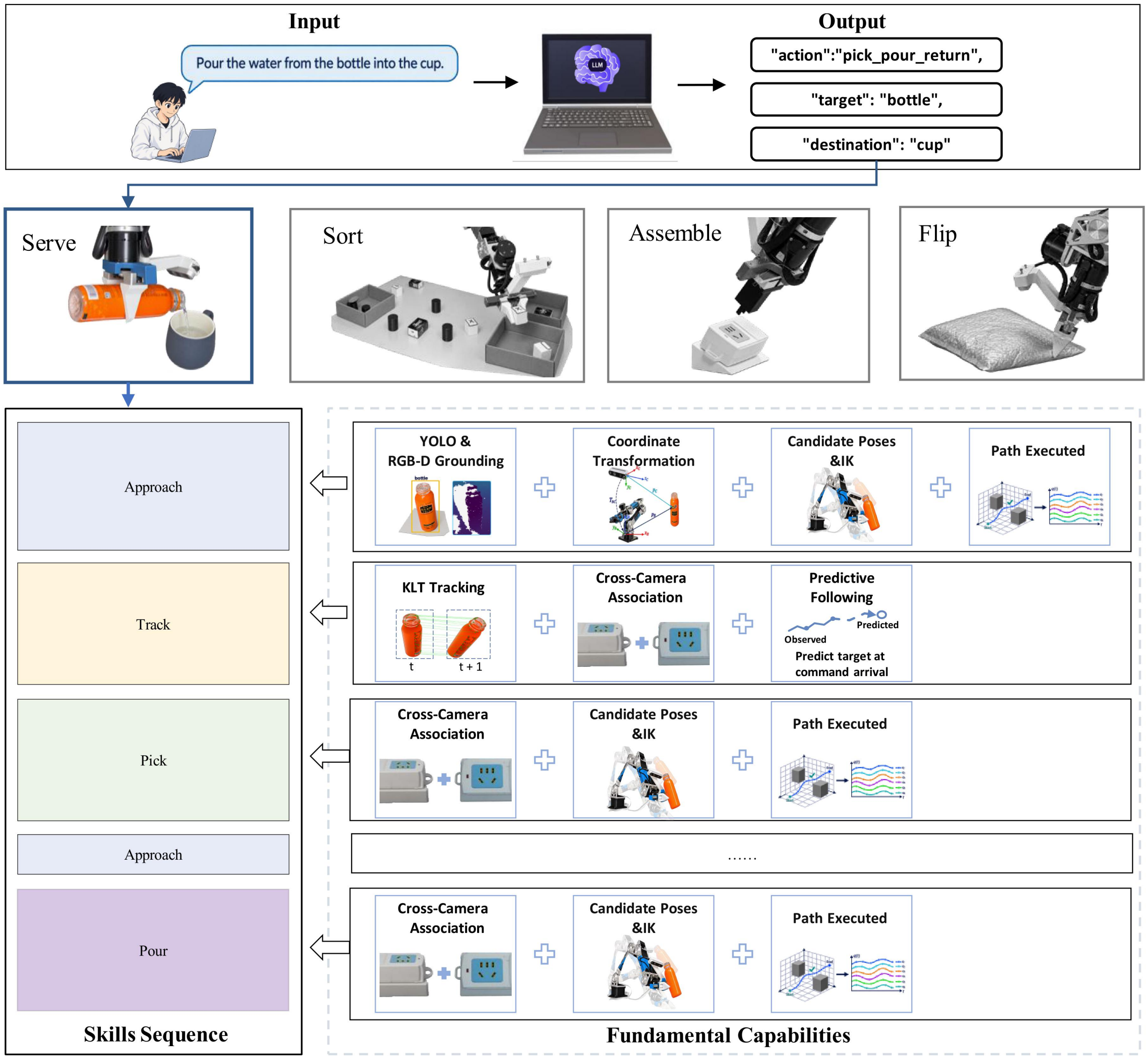

**Fig. 2 A Serve request is executed as a sequence of shared skills, each composed of fundamental capabilities.** The instruction is parsed into a task specification and compiled into the skill sequence Approach, Track, Pick, Approach and Pour. Each skill calls fundamental capabilities shared by all four evaluated tasks.

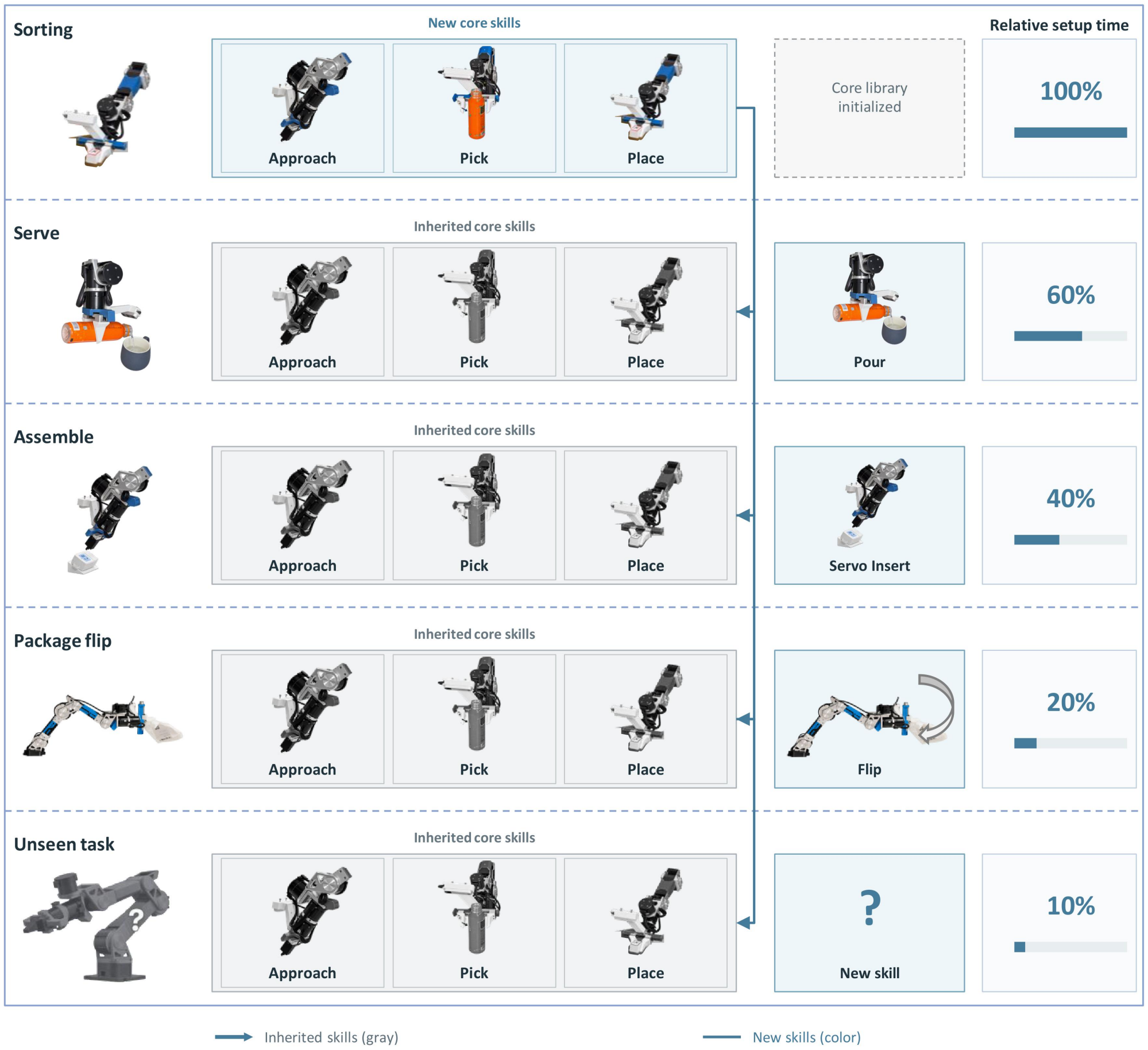

**Fig. 3 New tasks draw on the core skills and add only task-specific operations.** Each row lists the skills available to one task. Sorting establishes the core library (Approach, Pick, Place). Serve, Assemble and Flip draw on this library (grey) and add Pour, Servo Insert and Flip, respectively (colour). The bottom row represents a future task that adds one new skill. Bars on the right give relative setup time as illustrative values.

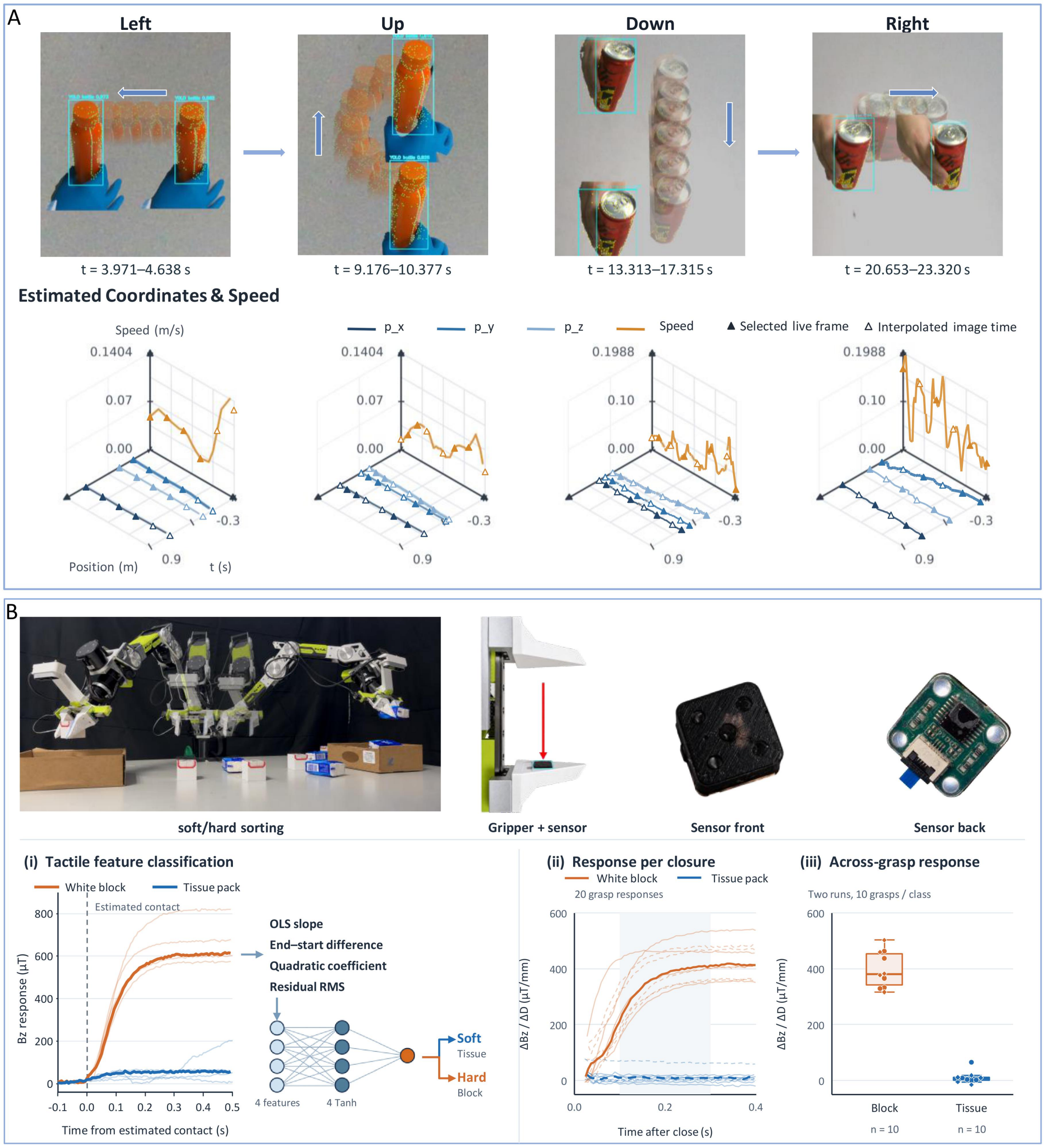

**Fig. 4 Visual tracking and tactile sensing supply the state on which skills act. A**, Tracking of a moving bottle and can, with estimated position and speed. **B**, Tactile sorting setup. **(i)** Magnetic response and soft/hard classifier. **(ii)** Closure-normalised responses. **(iii)** Per-grasp summary ($n = 10$ per class). Box plots: median, interquartile range (IQR), 1.5×IQR whiskers.

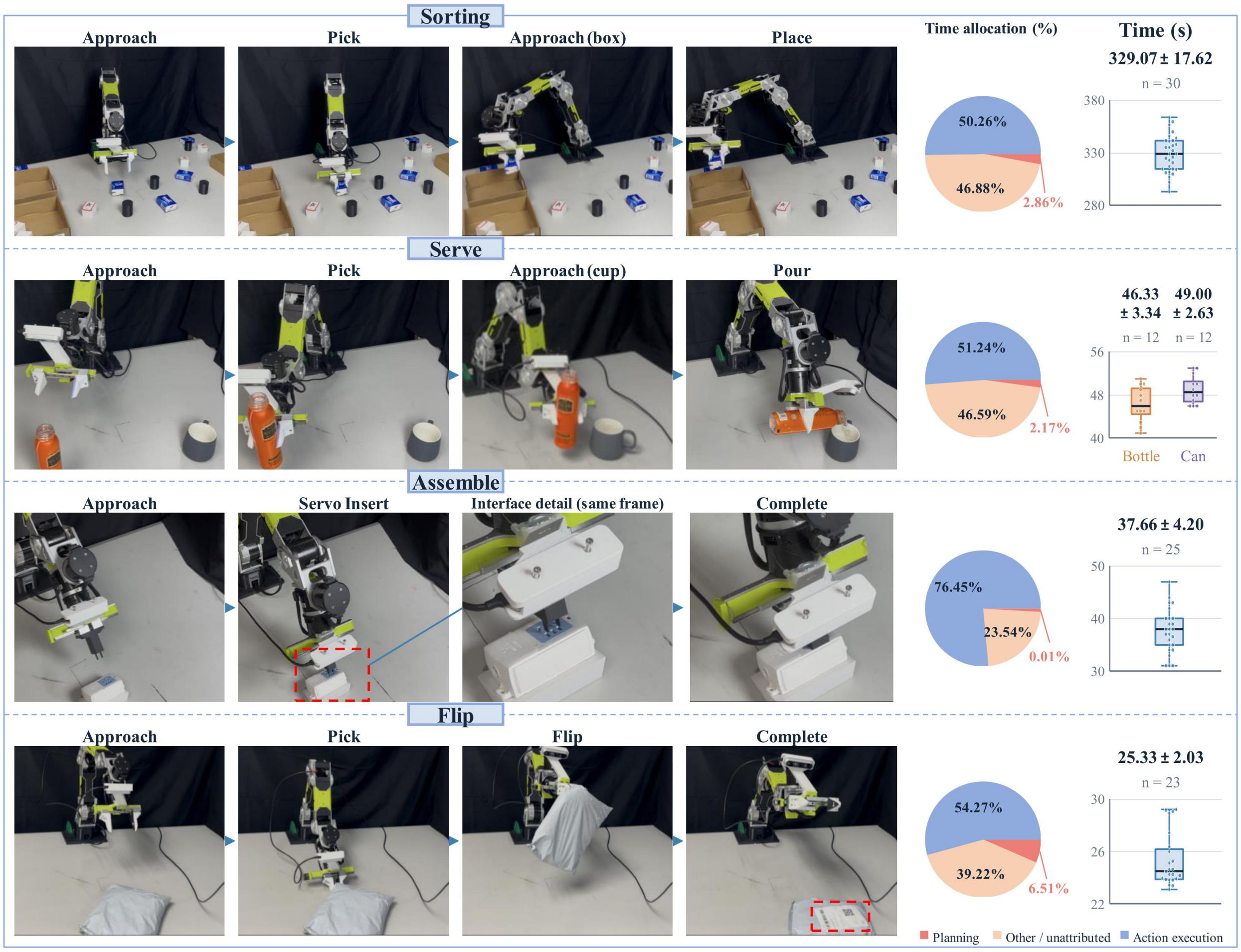

**Fig. 5 SHARE completes four real-robot tasks under randomised layouts, and planning accounts for only a small part of task time.** Rows, Sorting, Serve, Assemble and Flip. Left, representative stages of one trial; for Assemble, the inset enlarges the plug–socket interface in the same frame. Middle, composition of task time: planning took only 2.86%, 2.17%, 0.01% and 6.51% of task time in Sorting, Serve, Assemble and Flip, respectively, and action execution took 50.26%, 51.24%, 76.45% and 54.27%; the remainder was not separately attributed. Planning itself is fast, and task time is limited mainly by the execution of the robot hardware. Right, task duration from instruction input to task completion, for all 30 Sorting runs and for successful trials of Serve (bottle, $n = 12$; can, $n = 12$), Assemble ($n = 25$) and Flip ($n = 23$). Points, individual trials; box plots as in Fig. 4; values above, mean ± s.d. in seconds. Success rates and their 95% Wilson confidence intervals are given in the main text.